\documentclass[runningheads]{llncs}
\usepackage{cite}
\usepackage{amsmath,amssymb,amsfonts}
\usepackage{algorithm}
\usepackage{algorithmic}
\usepackage{graphicx}
\usepackage{textcomp}
\usepackage{xcolor}
\usepackage{hyperref}
\usepackage{xspace}
\usepackage{pgfplots}
\usepackage{tikz}
\usepackage{subcaption}

\newcommand{\reinit}{WMM-WR\xspace}    
\newcommand{\shuffle}{WMM-WS\xspace}    
\newcommand{\nn}{neural network\xspace}
\newcommand{\dnn}{deep \nn}
\newcommand{\dl}{deep learning\xspace}
\newcommand{\ai}{artificial intelligence\xspace}
\newcommand{\rl}{reinforcement learning\xspace}
\newcommand{\wmm}{WMM\xspace}

\begin{document}
\title{Stochastic Weight Matrix-based Regularization Methods for Deep Neural Networks\thanks{The research presented in this paper has been supported by the European Union, co-financed by the European Social Fund (EFOP-3.6.2-16-2017-00013, Thematic Fundamental Research Collaborations Grounding Innovation in Informatics and Infocommunications), by the BME-Artificial Intelligence FIKP grant of Ministry of Human Resources (BME FIKP-MI/SC), by Doctoral Research Scholarship of Ministry of Human Resources (\'UNKP-19-4-BME-189) in the scope of New National Excellence Program, by J\'anos Bolyai Research Scholarship of the Hungarian Academy of Sciences. We gratefully acknowledge the support of NVIDIA Corporation with the donation of the Titan Xp GPU used for this research.
Patrik Reizinger expresses his gratitude for the financial support of the Nokia Bell Labs Hungary.}}
\titlerunning{\wmm Regularization for Deep Neural Networks}
\author{Patrik Reizinger\inst{1} \and
B\'alint Gyires-T\'oth\inst{1}}
\authorrunning{P. Reizinger and B. Gyires-T\'oth}
\institute{{Budapest University of Technology and Economics}\\
Budapest, Hungary
\email{\{rpatrik96@sch, toth.b@tmit\}.bme.hu}}
\maketitle              % typeset the header of the contribution
\begin{abstract} The aim of this paper is to introduce two widely applicable regularization methods based on the direct modification of weight matrices. The first method, Weight Reinitialization, utilizes a simplified Bayesian assumption with partially resetting a sparse subset of the parameters. The second one, Weight Shuffling, introduces an entropy- and weight distribution-invariant non-white noise to the parameters. The latter can also be interpreted as an ensemble approach. The proposed methods are evaluated on benchmark datasets, such as MNIST, CIFAR-10 or the JSB Chorales database, and also on time series modeling tasks. We report gains both regarding performance and entropy of the analyzed networks. We also made our code available as a GitHub repository\footnote{\url{https://github.com/rpatrik96/lod-wmm-2019}}.

\keywords{Deep learning  \and Generalization \and Regularization \and Weight matrix.}
\end{abstract}
\section{Introduction}
\label{sec:intro}
The importance of mitigating overfitting, i.e., the reduction of the generalization ability of a \dnn, has gained more importance with the broadening of the spectrum of using \dl for solving real-world problems. This is mainly due to the fact that \ai has become a tool widely used in the academia and industry, and in several cases, no feasible way exists to collect enough or perfectly representative data.
The existing pool of methods is rather numerous, nevertheless, new approaches can bring additional advantages. Before introducing our proposal, it is worth briefly summarizing the aspects of the most popular regularization methods.
Traditional constraints on the parameters (like L1 and L2 norms) add a prior to the weight matrix distribution, so the method takes its effect not directly on the weights, as the methods described in \autoref{sec:proposal}. Novel results\cite{Masters2018} propose the usage of smaller batch sizes; more robust training and improved generalization performance is reported - while the variance of the minibatch, in general, tends to be higher.
Dropout\cite{dropout} and its generalization, DropConnect\cite{dropconnect}, are approaching the effect of ensemble networks by partially blocking different parts of the error propagation in the computational graph throughout the training. Batch Normalization\cite{batchnorm} affects the gradients, by taking a normalization step improvements are reported both regarding accuracy and rate of convergence.

Another approach of regularization, stemming from the scarceness of available data, is the augmentation of the dataset. For this purpose, data transformations (e.g. random cropping, rotations for images) are applied or noise is added, thus providing more samples with slight differences. However, in the case of working with, e.g. time series or natural language, besides noise, it is rather difficult to find the equivalent transformations which are extensively used in the case of computer vision for data augmentation. For time series classification, there exist methods of similar nature, such as Window Warping\cite{time_augment}, which scales a randomly selected slice of a signal.
A promising result regarding noise as means of regularization for \rl is given in \cite{Plappert2017}, the use of additive noise in the parameter space contributed to significant improvements in several tasks. For recurrent networks, another type of regularization method has turned out to be successful: Zoneout\cite{zoneout} can be considered as a variant of Dropout for recurrent neural networks such as LSTM (Long Short-Term Memory, \cite{lstm}). The main difference compared to Dropout is that Zoneout maintains at least the identity - i.e., instead of dropping specific cells, the activation values from the last step are used in the graph -, thus preserving information flow in the recurrent connections.

This paper is organized as follows: \autoref{sec:motivation} briefly reviews the motivation for the proposed methods, \autoref{sec:proposal} includes the detailed overview of the proposed algorithms, the experimental setup is described in \autoref{sec:experiment}, while the results are introduced in \autoref{sec:results} and discussed in \autoref{sec:conclusions}.

\section{Motivation}
\label{sec:motivation}
Considering the most widely-used approaches and the possibilities, we decided to investigate more thoroughly the direct effects of regularization applied to weight matrices. We refer to all of the tunable parameters of a layer (excluding hyper-parameters) as weight or weight matrix in this paper. While doing so, we also considered that in contrast to the real-world or artificial data, parameters do not possess such a high degree of internal error-resilience. I.e., an improperly designed method can lead to significant degradation of the training, even to complete failure. As weights capture information about the underlying stochastic process in a compact way, this is something which should be considered. Thus, such schemes are needed which do not cause significant parameter degradation - if degradation occurs, it should be corrected before triggering early stopping. This means, the intervention should be local and the change should be small.

Given these considerations and constraints, our goal is to design robust, generally applicable regularization methods based on the direct modification of the weight matrices.

\section{Proposed Methods}
\label{sec:proposal}
In this section, we describe both proposed methods, collectively named \wmm (Weight Matrix Modification). Each of them modifies weights explicitly, but the two approaches are different. \reinit (Weight Reinitialization) can be considered as a way of expressing distrust of backpropagation by discarding some of its parameter updates, while \shuffle (Weight Shuffling) is a local approach, which constrains its intervention based on an assumption regarding the autocorrelation function of a specific weight matrix.

In this section $W$ denotes a weight matrix, $p$ a probability value, $\mathcal{L}$ the set of affected layers and $c$ a constant (called \emph{coverage}). Thus, $p$ determines the probability of carrying out \reinit or \shuffle and $c$ specifies the size of the window where the method takes effect. Weight matrices are considered two dimensional as it is the general case (for 2D CNNs - Convolutional Neural Networks -, we consider the 2D filters separately -- not the filter bank as a whole). Nevertheless, it is done without loss of generalization, hence this property is not exploited in the proposed methods. Please note that both methods are concerning the weight matrices only and not the bias vectors, because our evaluations showed that including the bias cannot bring further advantages.

\subsection{Weight Reinitialization (\reinit)}
\label{subsec:reinit}
As mentioned in \autoref{sec:motivation}, the initial weight distribution can be crucial for good results. Like the weight initialization scheme of \cite{glorotweights}, or the common choice of
\begin{align}
    W &\sim U \left[-\dfrac{1}{\sqrt{n_i}}, \dfrac{1}{\sqrt{n_i}}\right], \label{eq:uniform_weights}
\end{align}
where $n_i$ is the number of columns of the weight matrix $W$, and $U$ denotes a uniform distribution. From an information theoretical point of view, using uniformly distributed weights corresponds to the principle of maximal entropy, thus a high degree of uncertainty. However, the cited paper from Glorot and Bengio\cite{glorotweights}, which can improve the convergence with normalizing the gradients, also uses a uniform parameter distribution.

These considerations provide us means for proposing a regularization method with a Bayesian approach. It is common in estimation theory to combine different probability distributions, often weighted with the inverse covariance matrix. This approach can also be used for weight matrices with a slight modification, which is needed to constrain the amount of change to maintain stability during training. For this, we apply a sparse mask. Since we only modify a small amount of the weights, it seemed not feasible to compute the statistics of the whole layer. Different sparse masks are generated in each step, which is also not possible for all combinations without significant speed degradation. Thus, the algorithm either preserves the actual weights or draws a sample from the initial uniform distribution.

\autoref{alg:reinit} describes the steps of \reinit. It is carried out based on a probabilistic scheme, which is controlled by the $p$ parameter. This approach reduces the probability to introduce a high level of degradation into the network, on the other hand, it also saves computational power, because the method will not be applied in every step. First, a mask $M$ is generated by sampling a uniform distribution $U$. Then a window position is selected randomly (its size is controlled by the $c$ parameter) -- which means that all items outside the window are cleared in $M$. After that the same $p$ is used as a threshold for ensuring sparsity, thus clearing the majority of the elements within the window. $W^{*}$ denotes the temporary matrix used for storing the samples from the distribution $U_w$, the distribution used for initializing the weights at the start of the training. Based on the mask values (which remained within the window after thresholding with $p$), the elements of the original weight matrix $W_i$ get reinitialized. Generating both $M$ and $U_w$ is linear in the number of the weights, as in the case of memory cost too.

In the case of overfitting, activations of a layer are distributed rather unevenly. As activation functions are in general injective, the distribution of the weights is uneven, too. By reinitializing a random subset of weights, if weights from the peaks of the weight distribution are chosen, overfitting can be reduced with doing a modification step towards maximal entropy, thus smoothing the histogram. On the other hand, reinitializing values near the mean would result in small changes only. If the mean of the weight matrix is not the same as the expected value of the trained weight matrix, then reinitializing a subset of weights expresses our distrust and is opting for maximal entropy, thus for uniform distribution. Independent of the value of the mean of the whole layer, the reinitialized subset itself has an expected value of zero (\autoref{eq:uniform_weights}), thus the method will not result in a significant change, meaning reduced sensitivity. An interesting side-effect of \reinit is the fact that it reduces the Kullback-Leibler divergence between the actual distribution of the weights and the initial one. The reason for this is that the same distribution is used for \reinit as for initializing the layers. The method increases the entropy (at least momentarily, as a uniform distribution is used), which effect will be investigated in \autoref{sec:results}.

\begin{algorithm}[t]
    \caption{Algorithm for \reinit}\label{alg:reinit}
    \begin{algorithmic}
        \REQUIRE $\mathcal{L}, p, c$
        \FORALL{$W_i \in \mathcal{L}$}
            \IF{$p > U\left[0;1\right)$}
                \STATE $M \gets U\left[0;1\right)$
                \STATE $M \gets window\left(M, c\right)$
                \STATE $M \gets threshold\left(M, p\right)$
                \STATE $W^{*} \gets U_w$
                \STATE $W_i\left[M_{jk} == 1\right] = W^{*}$
            \ENDIF
        \ENDFOR
    \end{algorithmic}
\end{algorithm}

\subsection{Weight Shuffling (\shuffle)}
\label{subsec:shuffle}
As already mentioned in \autoref{sec:intro}, noise has a wide range of applicability from the dataset to the weights. Instead of using Gaussian white noise, we consider using non-white, correlated noise directly for a neighborhood of weights. We refer to a neighborhood of weights as a rectangular window (determined by the parameter $c$) with dimensions comparably smaller than the size of the weight matrix itself.
Because the weights are intended to represent the characteristics of the underlying stochastic process, they should be correlated with each other (i.e. white noise itself is not appropriate for capturing the underlying process). Consequently, by selecting a continuous subset of the weights, the correlation should still hold, hence the process to be modeled stays the same. Shuffling them within the bounds of a local window can be interpreted as using correlated noise for that given subset. Interestingly, this transformation is invariant regarding both the weight distribution and the entropy of the matrix -- the latter of which will be investigated in \autoref{sec:results}.

A more expressive way is to think about \shuffle as a type of a 'parameter ensemble', i.e., weights in a neighborhood are forced to fit well a range of input locations. As the ensemble approach, or the approximations of it - such as Dropout -, have been proven to be successful in practice due to the utilization (or simulation) of an averaging step, an improved degree of robustness is expected from \shuffle as well.

The steps of \shuffle are described in \autoref{alg:shuffle}. As in the case of \autoref{alg:reinit}, we have a probabilistic first step which stochastically decides whether \shuffle will be carried out or not (using parameter $p$). The mask $M$ is sampled from a Bernoulli-distribution $\mathcal{B}$ directly (as we need a binary mask), thus the density of the mask is independent of the parameter $p$. For the selection of the local window, $c$ is used as for \reinit. Having generated the mask, the weights indexed by $M$ are shuffled for each weight matrix of the layer subset $\mathcal{L}$ (filters of convolutional and gates of recurrent layers are handled independently from each other). The complexity of the algorithm is determined with the shuffle operator, which scales linearly in the number of the weights while selecting the local window has a constant computational cost.

\begin{algorithm}[t]
    \caption{Algorithm for \shuffle}\label{alg:shuffle}
    \begin{algorithmic}
        \REQUIRE $\mathcal{L}, p, c$
        \FORALL{$W_i \in \mathcal{L}$}
            \IF{$p > U\left[0;1\right)$}
                \STATE $M \gets \mathcal{B}\left[0;1\right]$
                \STATE $M \gets window\left(M, c\right)$
                \STATE $shuffle\left(W_i\left[M\right]\right)$
            \ENDIF
        \ENDFOR
    \end{algorithmic}
\end{algorithm}

\section{Experimental setup}
\label{sec:experiment}
Due to the fact that \wmm intervenes into the training loop on the lower levels in the software stack, we have searched for a framework which provides such a low-level interface. PyTorch\cite{Paszke2017} was chosen to be used for the implementation because it makes possible to modify fine details of the training process. We utilized Ignite\cite{ignite}, a high-level library for PyTorch, which was built by the community to accelerate the implementation of \dl models by providing a collection of frequently used routines.

\begin{figure}[t]
    \centering
\includegraphics[width=.9\columnwidth,height=.9\textheight,keepaspectratio]{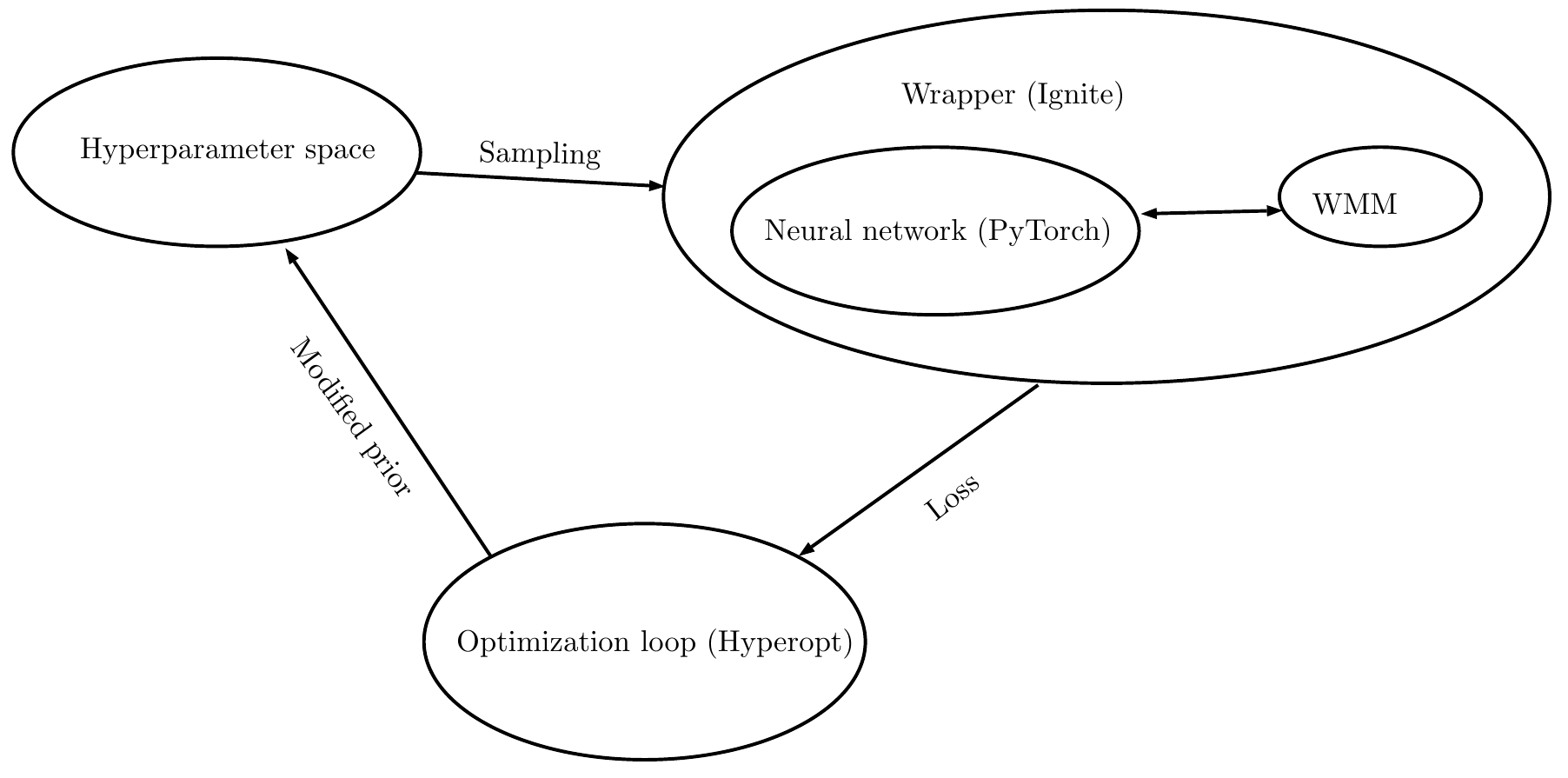}
    \caption{Structure of the framework for \wmm}
    \label{fig:sw_struct}
\end{figure}

The framework of the proposed methods is schematically depicted in \autoref{fig:sw_struct}: Ignite provides the wrapper which merges the PyTorch model with \wmm utilizing its callback interface. Due to the fact that while designing \wmm neither architecture- nor task-specific assumptions were made, our approach provides great flexibility, which will be elaborated shortly. For hyper-optimization, a Bayesian approach (tree of Parzen estimators) was used, based on a method shown in \cite{tpebergstra}.

Regarding the parameters of \wmm, the following intervals were chosen. For $p$ and $c$ (coverage) a log-uniform distribution was selected to avoid a big value for both parameters at the same time, the intervals were $\left[0.05;0.4\right]$ and $\left[0.03;0.35\right]$, respectively. Each layer was affected by \shuffle or \reinit with an equal probability.

To prove the general applicability of the proposed methods, we set up a testing environment with different model architectures and datasets. Tests were carried out with fully connected, convolutional and LSTM (Long Short-Term Memory, \cite{lstm}) layers. In case of convolutional layers, regardless of the filterbank representation of the kernels, \wmm operations were, of course, restricted to operate only within one filter. For LSTM the effect of \wmm on different gates was investigated, the same holds for GRU (Gated Recurrent Unit, \cite{Cho2014}) as well.

For evaluation, several datasets were investigated. The MNIST\cite{LeCun1998} dataset and its sequential (row-by-row) variant were split up into 55,000-5,000-10,000 samples (28x28 pixels, 10 classes, for each handwritten digit) for training, validation, and test purposes. The networks consisted of 2 convolutional and 2 fully connected or 2 LSTM and 1 fully connected layers for the traditional and sequential approaches, respectively. For CIFAR-10\cite{Krizhevsky2009} the same splitting policy was used (the samples are 32x32x3, 10 classes), in that case, the network consisted of 2 convolutional (with a max pooling between them) and 3 fully connected layers. The JSB Chorales\cite{jsb_chorales} task was organized into 229-76-77 chorales (with a window length of 25, 88 classes) for training, validation, and test purposed. Although being a multilabel problem a one-layer LSTM was sufficient. 
Besides classification, regression problems were also investigated, for that purpose an artificially generated (synthetic) dataset was used, using also a 55,000-5,000-10,000 data split (window size 50). The synthetic time series were fed into the \nn both with and without colored noise, and consisted of the linear combination of sinusoidal, exponential and logarithmic functions. Our intent with this choice - which is somewhat similar to that of the Phased LSTM\cite{phased_lstm} paper - was, to begin with simpler datasets and proceed to more complex ones. For both variants of the regression task, a network of 2 LSTM and 1 fully connected layers was used.

\section{Results and discussion}
\label{sec:results}
During the analysis of the results, several factors were considered. The three most important factors were performance (accuracy/MSE (Mean Squared Error)), entropy and robustness.

\begin{figure}[!ht]
    \centering
    \begin{subfigure}{0.8\textwidth}
        \centering    \includegraphics[width=\linewidth,height=\textheight,keepaspectratio]{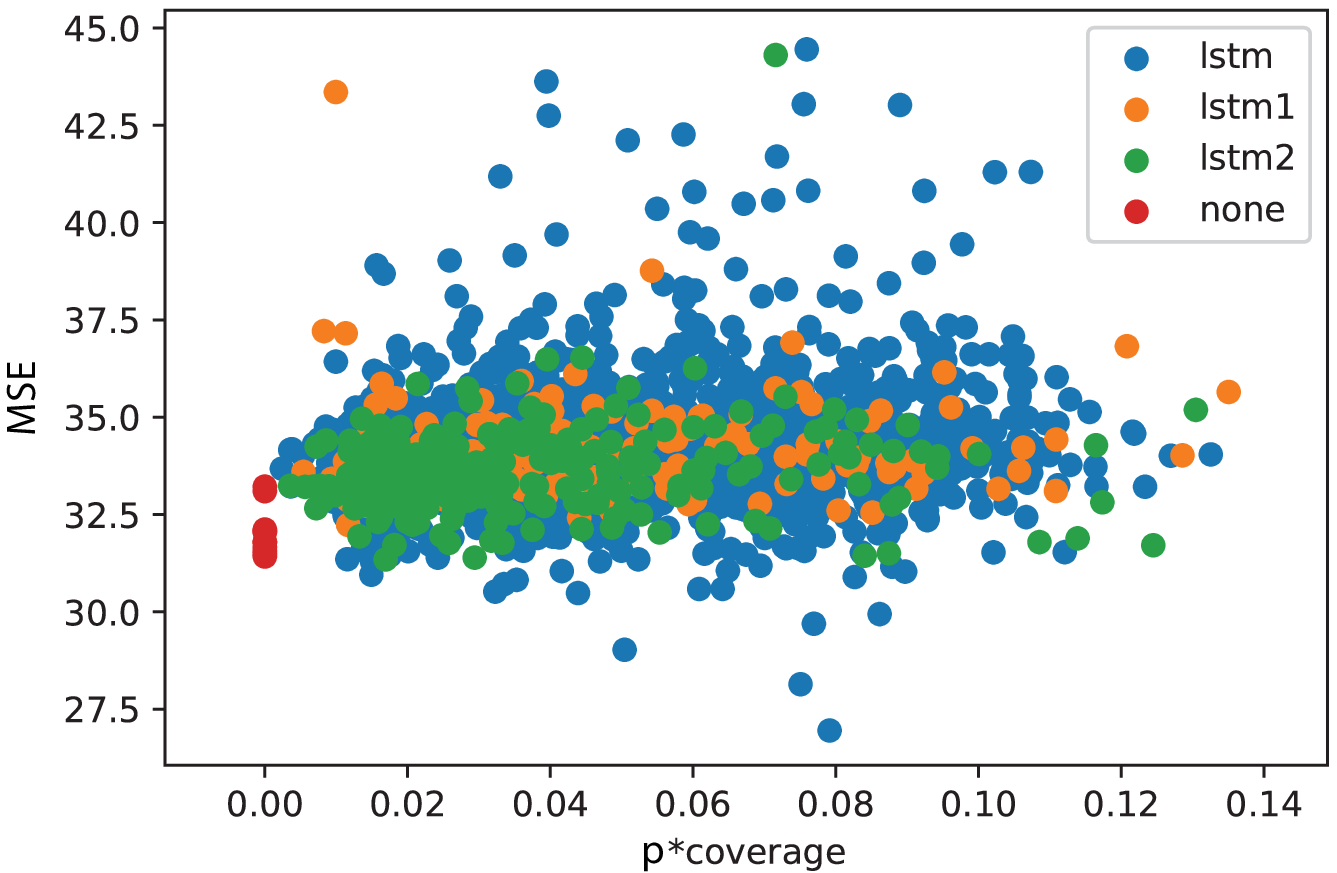}
        \caption{\reinit}
        \label{fig:mse_p_cov_sin-noise_r}
    \end{subfigure}
    \begin{subfigure}{0.8\textwidth}
        \centering
        \includegraphics[width=\linewidth,height=\textheight,keepaspectratio]{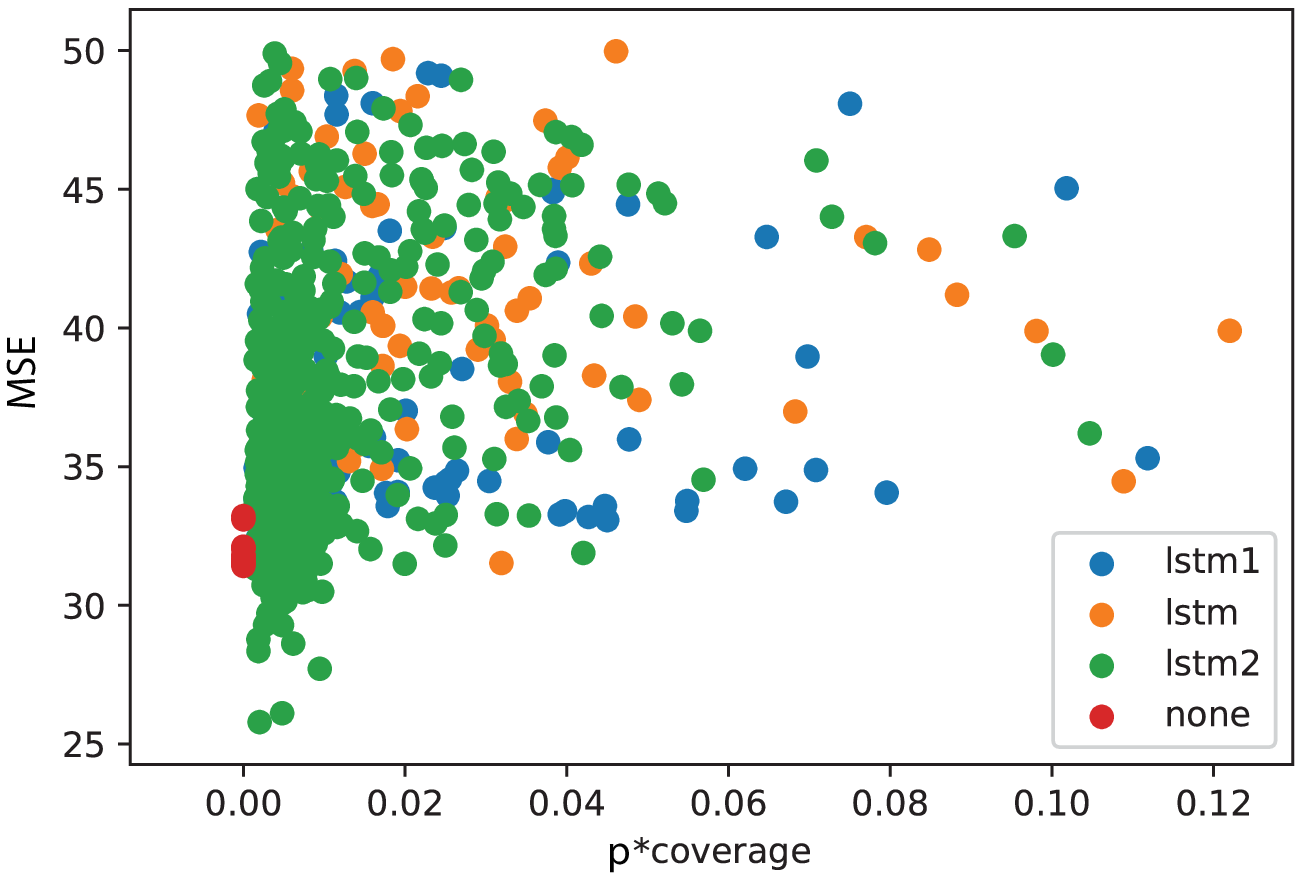}
        \caption{\shuffle}
        \label{fig:mse_p_cov_sin-noise_s}
    \end{subfigure}
    \caption{Illustration of the robustness examination for the synthetic dataset (with noise)}
    \label{fig:param_sin-noise}
\end{figure}

Our analysis intended to identify the tasks where the proposed approach turns out to be advantageous. Nonetheless, we could even make useful conclusions with datasets such as MNIST, where state-of-the-art results are near perfect, using it as a testbed for algorithm stability. For each dataset and each of the proposed methods, we have conducted more than $2000$ experiments. The aim of which was to evaluate the behavior of both methods empirically, mainly for describing parameter sensitivity (see \autoref{fig:param_sin-noise} and its discussion below).

\pgfplotstableread[row sep=\\,col sep=&]{
    interval            & b  & r   & s \\
    MNIST               & 1.0       & 1.00       & 1.00  \\
    Sequential MNIST    & 1.0       & 1.00       & 1.00 \\
    JSB Chorales        & 1.0       & 1.00       & 1.00 \\
    CIFAR-10            & 1.0       & 1.07      & 1.03 \\
    Synthetic           & 1.0       & 0.95      & 0.86 \\
    Synthetic + Noise     & 1.0       & 1.12      & 1.16 \\
    }\performance
\begin{figure}[t]
    \centering
    \begin{tikzpicture}
        \begin{axis}[
                ybar,
                bar width=.25cm,
                width=\columnwidth,
                height=.5\textwidth,
                legend style={at={(0.5,1.15)},
                    anchor=north},
                symbolic x coords={MNIST, Sequential MNIST, JSB Chorales, CIFAR-10,Synthetic,Synthetic + Noise},
                xticklabel style={rotate=90,text width=1.5cm, align=center},
                xtick=data,
                xlabel={Dataset},
                x label style={at={(axis description cs:0.5,-0.3)},anchor=north},
                nodes near coords,
                every node near coord/.append style={rotate=90, at={(axis description cs:0.06,0.1)},anchor=north},
                nodes near coords align={vertical},
                ymin=0.7,ymax=1.3,
                ylabel={Performance relative to the reference model (\%)},
            ]
            \addplot table[x=interval,y=b]{\performance};
            \addplot table[x=interval,y=r]{\performance};
            \addplot table[x=interval,y=s]{\performance};
            \legend{Reference (best), \reinit (average of top 5), \shuffle (average of top 5)}
        \end{axis}
    \end{tikzpicture}
    \caption{The effect of \wmm on performance metrics (higher values are better)}
    \label{fig:perf}
\end{figure}
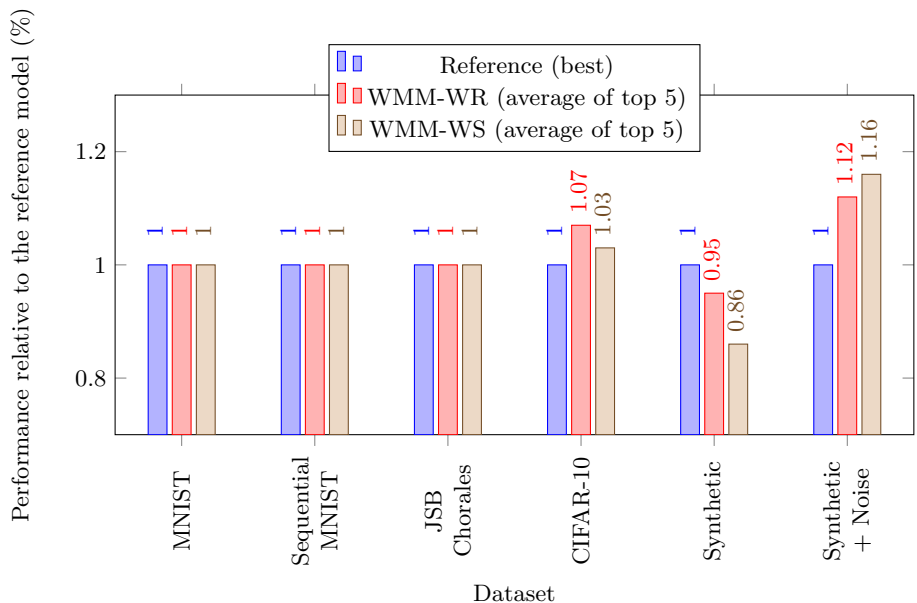

To clearly identify the effects of \wmm and to prevent overfitting originating from the simple nature of some datasets, the networks used for evaluation were not very deep (maximum depth was 5 layers).
During the evaluation, a reference model (for each task, respectively) was constructed from the best result using L2-regularization, Batch Normalization\cite{batchnorm} or Dropout\cite{dropout}, which were chosen as they are currently among the most successful and most widely-used regularization techniques.

The figures included in this section are comparing different aspects of the networks. We compared the average of the top 5 best results utilizing \wmm to the best results of other regularization techniques. The effect of the proposed methods regarding performance metrics is depicted in \autoref{fig:perf} (higher is better), we converted accuracy and MSE to be able to display them on the same scale (it displays relative change).

Regarding performance, the change is not significant for MNIST (traditional and sequential approach) and JSB Chorales (below $0.1\%$). In the case of the synthetic dataset, performance suffers even from degradation. On the other hand, for CIFAR-10 \reinit brought a performance boost of $7\%$, while \shuffle one of $3\%$. The biggest performance increase (over $10\%$) was experienced in the case of the synthetic sequential dataset with colored noise.

For the tasks utilizing recurrent networks (sequential MNIST, JSB Chorales, synthetic data) the best results were achieved by different choices of the LSTM gates. In case of \reinit the modification of the weights of the forget gate was proven to be the best for the sequential MNIST task, while for the other three problems the cell gate was the best choice. For \shuffle the output gate was preferred (synthetic + JSB Chorales) -- the sequential MNIST task resulted in modifying the weights of the last, fully connected layer.

\pgfplotstableread[row sep=\\,col sep=&]{
    interval            & b  & r   & s \\
    MNIST               & 1.0       & 0.97       & 1.01  \\
    Sequential MNIST    & 1.0       & 0.91       & 0.98 \\
    JSB Chorales        & 1.0       & 1.02       & 0.52 \\
    CIFAR-10            & 1.0       & 0.85      & 0.92 \\
    Synthetic           & 1.0       & 0.82      & 0.51 \\
    Synthetic + Noise     & 1.0       & 0.65      & 0.57 \\
    }\entropy
\begin{figure}[t]
    \centering
    \begin{tikzpicture}
        \begin{axis}[
                ybar,
                bar width=.25cm,
                width=\columnwidth,
                height=.5\textwidth,
                legend style={at={(0.5,1.25)},
                    anchor=north},
                symbolic x coords={MNIST, Sequential MNIST, JSB Chorales, CIFAR-10,Synthetic,Synthetic + Noise},
                xticklabel style={rotate=90,text width=1.5cm, align=center},
                xtick=data,
                xlabel={Dataset},
                x label style={at={(axis description cs:0.5,-0.3)},anchor=north},
                nodes near coords,
                every node near coord/.append style={rotate=90, at={(axis description cs:0.06,0.1)},anchor=north},
                nodes near coords align={vertical},
                ymin=0.3,ymax=1.3,
                ylabel={Entropy relative to the reference model (\%)},
            ]
            \addplot table[x=interval,y=b]{\entropy};
            \addplot table[x=interval,y=r]{\entropy};
            \addplot table[x=interval,y=s]{\entropy};
            \legend{Reference (best), \reinit (average of top 5), \shuffle (average of top 5)}
        \end{axis}
    \end{tikzpicture}
    \caption{The effect of \wmm onto the entropy of neural networks (lower values are better)}
    \label{fig:entropy}
\end{figure}
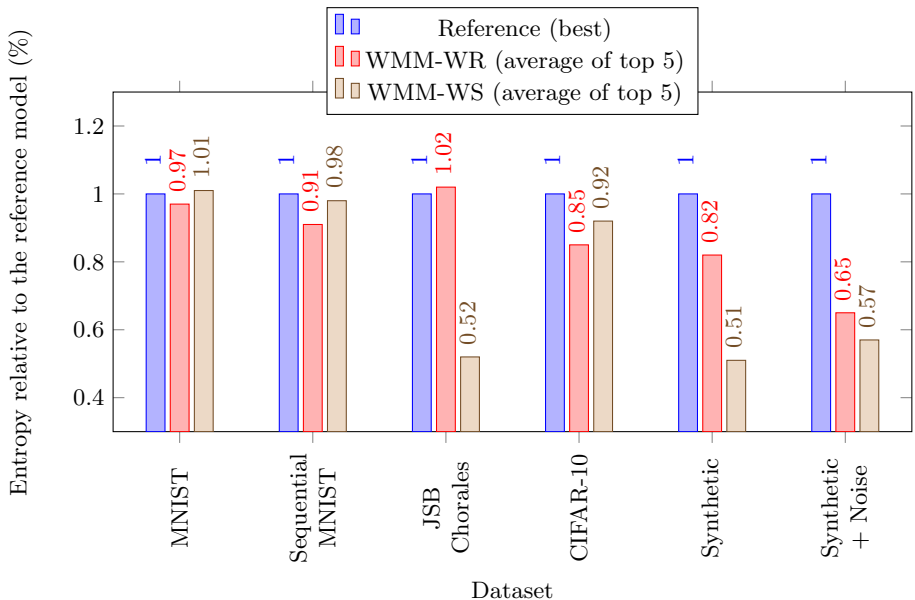

Partially due to the fact that the proposed methods have an effect onto the entropy of the networks -- \reinit is expected to increase (as the distribution used for reinitialization is uniform, thus has maximal entropy), while \shuffle to retain it -- so we decided to compare also entropy values for the different tasks. These results are shown in \autoref{fig:entropy} (lower is better), where we also compare to the respective reference models (which are the best using other regularization methods).

For MNIST the change is not too significant, but in general, \wmm was able to compress the networks by $2-9\%$, except \shuffle for the not sequential task. For CIFAR-10, the compression is more significant, both methods rank in the same range.

In the case of the only multilabel dataset, JSB Chorales, the results were rather astonishing: while \reinit has not brought a significant change,\\ \shuffle compressed the network by almost a factor of two. This superiority of \shuffle compared to \reinit regarding data compression is also characteristic of both variants of the synthetic dataset. Nevertheless, in the latter case \reinit also reduced entropy for the synthetic data, but there is a difference of $27\%$ between the noiseless and noisy cases.

\begin{figure}[ht]
        \centering
        \includegraphics[width=.8\columnwidth,height=.8\textheight,keepaspectratio]{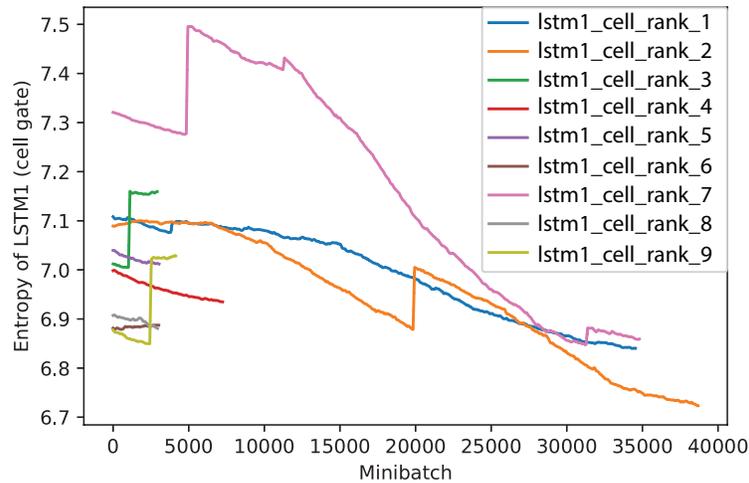}
        \caption{Temporal entropy change during \reinit (JSB Chorales, lower values are better)}
        \label{fig:entropy_jsb-chorales_cell_r}
\end{figure}

Generally speaking, \shuffle with its entropy-invariant approach was able to reduce entropy more than \reinit. Despite using a probability distribution according to the principle of maximal entropy, \reinit is also able to compress the networks. The analysis of the temporal change of entropy in the case of \reinit is shown in \autoref{fig:entropy_jsb-chorales_cell_r} using the best results, where the legend shows the $\mathcal{L}$ parameters of \autoref{alg:reinit}, also including the gates of the LSTM layer and the ranking according to MSE (the smaller the number, the lower the MSE). Although the entropy momentarily increases due to \reinit, in the end, it can help to compress the network, as shown in \autoref{fig:entropy_jsb-chorales_cell_r}.

The robustness of the proposed methods is also a crucial factor. \autoref{fig:param_sin-noise} shows the distribution of MSE w.r.t $p*c$ ($c$ is the coverage parameter) during Bayesian hyper-optimization. The $\mathcal{L}$ parameter is color-coded in the figures, \emph{none} denotes models without \wmm (i.e. the reference). In the case of \reinit(\autoref{fig:mse_p_cov_sin-noise_r}), the plot has a horizontally elongated form, which means smaller sensitivity to parameter change (the less elongation in the y-direction, the better). Although using \reinit imposes a higher variance due to its stochastic nature (the scatter plot displays the results of the whole hyper-optimization loop, containing also the not that good results at the beginning), it is clear to see that better results can be achieved with it.
A rather unexpected observation was, that in the case of \shuffle(\autoref{fig:mse_p_cov_sin-noise_s}) the bounding ellipse is rotated with $90^{\circ}$ compared to \autoref{fig:mse_p_cov_sin-noise_r}, thus showing higher stochastic sensitivity - nevertheless, the Bayesian optimization loop resulted in several good parameter sets.

\section{Conclusions}
\label{sec:conclusions}
In this paper two, weight matrix-based regularization methods were proposed and investigated for \dnn applications. The main advantages of both \reinit and \shuffle are their architecture- and task-independence. 
Our evaluations resulted in the following conclusions: both methods are capable of reducing overfitting, first of all in the case of noisy data, but also for CIFAR-10 better results were reported for the networks used throughout testing. On the other hand, \reinit and \shuffle also have shown their potential concerning model compression. The capabilities of \wmm regarding noise reduction will be investigated more thoroughly in the future, we also intend to introduce an adaptive way (i.e. based on the statistics of the weight matrix) for selecting the by \wmm affected areas.

\bibliographystyle{splncs04}
\bibliography{WMM}

\begin{thebibliography}{10}
\providecommand{\url}[1]{\texttt{#1}}
\providecommand{\urlprefix}{URL }
\providecommand{\doi}[1]{https://doi.org/#1}

\bibitem{ignite}
Ignite. \url{https://github.com/pytorch/ignite} (2019)

\bibitem{tpebergstra}
Bergstra, J., Bardenet, R., Bengio, Y., K{\'{e}}gl, B.: {Algorithms for
  Hyper-Parameter Optimization}. Advances in Neural Information Processing
  Systems (NIPS) pp. 2546--2554 (2011). \doi{2012arXiv1206.2944S}

\bibitem{Cho2014}
Cho, K., {Van Merri{\"e}nboer}, B., Gulcehre, C., Bahdanau, D., Bougares, F.,
  Schwenk, H., Bengio, Y.: {Learning Phrase Representations Using RNN
  Encoder-Decoder for Statistical Machine Translation}. arXiv  (2014).
  \doi{10.1074/jbc.M608066200}

\bibitem{glorotweights}
Glorot, X., Bengio, Y.: {Understanding the Difficulty of Training Deep
  Feedforward Neural Networks}. Pmlr  \textbf{9},  249--256 (2010).
  \doi{10.1.1.207.2059}

\bibitem{time_augment}
Guennec, A.L., Malinowski, S., Tavenard, R., Guennec, A.L., Malinowski, S.,
  Tavenard, R., Augmentation, D., Guennec, A.L., Malinowski, S., Tavenard, R.:
  {Data Augmentation for Time Series Classification Using Convolutional Neural
  Networks}. In: ECML/PKDD Workshop on Advanced Analytics and Learning on
  Temporal Data (2016)

\bibitem{lstm}
Hochreiter, S., Schmidhuber, J.: {Long Short-Term Memory}. Neural Computation
  \textbf{9}(8),  1735--1780 (1997). \doi{10.1162/neco.1997.9.8.1735}

\bibitem{batchnorm}
Ioffe, S., Szegedy, C.: {Batch Normalization : Accelerating Deep Network
  Training by Reducing Internal Covariate Shift}. arXiv  (2015)

\bibitem{Krizhevsky2009}
Krizhevsky, A., Hinton, G.: {Learning Multiple Layers of Features from Tiny
  Images}. Tech. rep., Citeseer (2009). \doi{10.1.1.222.9220}

\bibitem{zoneout}
Krueger, D., Maharaj, T., Kram{\'{a}}r, J., Pezeshki, M., Ballas, N., Ke, N.R.,
  Goyal, A., Bengio, Y., Courville, A., Pal, C.: {Zoneout: Regularizing RNNs by
  Randomly Preserving Hidden Activations}. arXiv pp. 1--11 (2016).
  \doi{10.1227/01.NEU.0000210260.55124.A4},
  \url{http://arxiv.org/abs/1606.01305}

\bibitem{LeCun1998}
LeCun, Y., Bottou, L., Bengio, Y., Haffner, P.: {Gradient-based Learning
  Applied to Document Recognition}. Proceedings of the IEEE  \textbf{86}(11),
  2278--2323 (1998). \doi{10.1109/5.726791}

\bibitem{Masters2018}
Masters, D., Luschi, C.: {Revisiting Small Batch Training for Deep Neural
  Networks}. arXiv pp. 1--18 (2018). \doi{10.1016/j.biortech.2007.04.007},
  \url{http://arxiv.org/abs/1804.07612}

\bibitem{phased_lstm}
Neil, D., Pfeiffer, M., Liu, S.C.: Phased lstm: Accelerating recurrent network
  training for long or event-based sequences. In: Lee, D.D., Sugiyama, M.,
  Luxburg, U.V., Guyon, I., Garnett, R. (eds.) Advances in Neural Information
  Processing Systems 29, pp. 3882--3890. Curran Associates, Inc. (2016)

\bibitem{Paszke2017}
Paszke, A., Chanan, G., Lin, Z., Gross, S., Yang, E., Antiga, L., Devito, Z.:
  {Automatic Differentiation in PyTorch}. Advances in Neural Information
  Processing Systems 30 (Nips), ~1--4 (2017)

\bibitem{Plappert2017}
Plappert, M., Houthooft, R., Dhariwal, P., Sidor, S., Chen, R.Y., Chen, X.,
  Asfour, T., Abbeel, P., Andrychowicz, M.: {Parameter Space Noise for
  Exploration}. arXiv pp. 1--18 (2017), \url{http://arxiv.org/abs/1706.01905}

\bibitem{jsb_chorales}
Ra{\'s}, Z., Wieczorkowska, A.: {Advances in Music Information Retrieval,
  Series: Studies in Computational Intelligence, Vol. 274}. Springer-Verlag,
  Berlin Heidelberg (2010)

\bibitem{dropout}
Srivastava, N., Hinton, G., Krizhevsky, A., Sutskever, I., Salakhutdinov, R.:
  {Dropout: A Simple Way to Prevent Neural Networks from Overfitting}. Journal
  of Machine Learning Research  \textbf{15},  1929--1958 (2014).
  \doi{10.1214/12-AOS1000}

\bibitem{dropconnect}
Wan, L., Zeiler, M., Zhang, S., LeCun, Y., Fergus, R.: {Regularization of
  Neural Networks Using DropConnect}. Icml (1),  109--111 (2013).
  \doi{10.1109/TPAMI.2017.2703082}

\end{thebibliography}

\end{document}